%% file: main.tex
\documentclass[conference,compsoc]{IEEEtran}
\IEEEoverridecommandlockouts
\input{ethos.tex}
\usepackage{cite}
\usepackage{amsmath,amssymb,amsfonts}
\usepackage{algorithmic}
\usepackage{graphicx}
\usepackage{textcomp}
\usepackage{xcolor}
\usepackage{array}
\usepackage{multirow}
\def\BibTeX{{\rm B\kern-.05em{\sc i\kern-.025em b}\kern-.08em
    T\kern-.1667em\lower.7ex\hbox{E}\kern-.125emX}}

\usepackage{hyperref}
\usepackage[hyphenbreaks]{breakurl}

\begin{document}

\title{Taking off the Rose-Tinted Glasses:\\A Critical Look at Adversarial ML Through the Lens of Evasion Attacks}

\author{
    \IEEEauthorblockN{Kevin Eykholt}
    \IEEEauthorblockA{\textit{IBM Research} \\ kheykholt@ibm.com
    }
    \and
    \IEEEauthorblockN{Farhan Ahmed}
    \IEEEauthorblockA{\textit{IBM Research} \\ Farhan.Ahmed@ibm.com
    }
    \and
    \IEEEauthorblockN{Pratik Vaishnavi}
    \IEEEauthorblockA{\textit{Stony Brook University} \\ pvaishnavi@cs.stonybrook.edu
    }
    \and
    \IEEEauthorblockN{Amir Rahmati}
    \IEEEauthorblockA{\textit{Stony Brook University} \\ amir@cs.stonybrook.edu
    }
}

\maketitle

\begin{abstract}
The vulnerability of machine learning models in adversarial scenarios has garnered significant interest in the academic community over the past decade, resulting in a myriad of attacks and defenses. However, while the community appears to be overtly successful in devising new attacks across new contexts, the development of defenses has stalled. After a decade of research, we appear no closer to securing AI applications beyond additional training. Despite a lack of effective mitigations, AI development and its incorporation into existing systems charge full speed ahead with the rise of generative AI and large language models. Will our ineffectiveness in developing solutions to adversarial threats further extend to these new technologies?

In this paper, we argue that overly permissive attack and overly restrictive defensive threat models have hampered defense development in the ML domain. Through the lens of adversarial evasion attacks against neural networks, we critically examine common attack assumptions, such as the ability to bypass any defense not explicitly built into the model. We argue that these flawed assumptions, seen as reasonable by the community based on paper acceptance, have encouraged the development of adversarial attacks that map poorly to real-world scenarios. In turn, new defenses evaluated against these very attacks are inadvertently required to be almost perfect and incorporated as part of the model. But do they need to? In practice, machine learning models are deployed as a small component of a larger system. We analyze adversarial machine learning from a system security perspective rather than an AI perspective and its implications for emerging AI paradigms.

\end{abstract}


\section{Introduction}
Large-scale interest in foundation models such as  ChatGPT has revitalized public interest in deploying semi- or fully-automated AI system pipelines. At the same time, trusting AI, especially for large-scale deployments, remains an ever-present issue. We've observed an increasing number of proposals to develop reporting schemes for AI across various trust pillars such as fairness, explainability, and security to increase AI transparency~\cite{factsheets, modelcards, systemcards}. These ``AI nutrition facts'' are intended to assist model consumers in understanding the risks associated with a model. In particular, the security of machine learning models has begun attracting industrial interest. In 2020, Microsoft released a survey highlighting industrial concern regarding the threat of adversarial attacks~\cite{9283867}. Motivated by such concerns, others have released knowledge bases like MITRE ATLAS and the AI Risk Database to systematize intelligence related to AI security~\cite{mitre:atlas, airisk_database}.

Initially of academic interest only, adversarial attacks, such as evasion and poisoning attacks, focused on adversely impacting a model's reliability and performance. These attacks, however, were designed with somewhat unrealistic assumptions, such as continuous-valued input data (\eg images) and unrestricted access to the model. As such, security evaluations of proposed defenses have also relied on similar assumptions. Notably, Athalye~\etal identified that most proposed evasion defenses rely on gradient obfuscation, which could be easily overcome with minimal effort by an attacker~\cite{athalye2018obfuscated}. Their work, among others, encouraged practitioners to thoroughly evaluate proposed defenses against \textit{adaptive attackers} with knowledge of the defense measures and the ability to freely query the model. Athalye~\etal remark that \textit{``It is not meaningful to restrict the computational power of an adversary artificially (\eg to fewer than several thousand attack iterations)''}.

The evaluation style proposed by Athalye~\etal uses the worst-case security scenario of an AI power system: direct and unrestricted access to a machine learning model to highlight potential weaknesses. This threat model effectively isolates the machine learning model from the system it powers. Consequently, most published defenses rely on direct modifications to the machine learning model to either filter out adversarial inputs or improve the model's performance on an adversarial distribution. For example, all of the 31 evasion defenses listed in the IBM Adversarial Robustness Toolbox~\cite{art2018}, an open-source Python library for evaluating AI security, fall into one of these two categories. In short, \textit{a potential adversarial defense must perfectly secure an isolated model in the worst case scenario, remain future-proof, and any defense that does not achieve this standard is considered unworthy of further study}.

We observe that at most major security conferences, there are at least a few papers that describe adversarial attacks against a specific domain, typically assuming this attack scenario. But are these attacks plausible? If a threat actor were to replicate the attack against a real system, one might ask:
\begin{enumerate}
    \item What are the prerequisites to implement such an attack?
    \item Does the execution of the attack against the model indicate a larger non-machine learning security issue in the system (\eg weak passwords, lack of monitoring, data leak)?
    \item Do we have existing solutions to address such issues?
\end{enumerate}
Such questions are important to consider as the development and publication of adversarial defenses are directly influenced by what the community considers ``reasonable threats'' to machine learning models. Specifically, these questions consider the end-to-end security of an AI system rather than just the security of the model. A cybersecurity attack against an AI system occurs across multiple phases and comprises a series of \textit{tactics}~\cite{mitre:atlas}, in which adversarial attacks such as evasion and poisoning are among middle-stage tactics. In this paper, we argue that the current community perspective of adversarial machine learning has been very lenient with adversarial attacks and harmful to the development of practical adversarial defenses. In particular, our paper makes the following contributions:
\begin{enumerate}
    \item We examine a recently published adversarial evasion attack to represent the current community perspective of reasonable attacks and identify potential feasibility issues. Specifically, we show that the strong assumptions made by the attack and others using the Athayle~\etal threat model are unreasonable for real systems, and if these assumptions were true, it would indicate a more significant security issue at play across the larger system.
    \item We expand our study to attacks ``in the wild'' and identify some common themes required for a successful attack.
    \item Based on our analysis, we identify that existing adversarial evasion attacks make simplifying assumptions that skip straight to the \textit{exploitation phase} 
    without considering the other necessary attack steps. This oversight can be attributed to the community's focus on defending the model rather than the system.
    These attack steps, if prevented, can hinder or stop an attack entirely. 
    \item We present one example of how traditional security techniques can be applied to mitigate adversarial evasion attacks. Reconnaissance is a necessary step for all evasion attacks, but due to the overly permissive attack threat model, most attacks do not focus on this aspect. As such, we measure the visibility of adversarial reconnaissance. 
    \item Given the visibility of current adversarial evasion attacks, we show how some simple detection techniques can prevent existing attacks before they even start. While these techniques may not necessarily apply to all machine learning applications, our study highlights how anomalous adversarial reconnaissance is compared to benign interactions with the model.
\end{enumerate}

Our literature survey suggests a lack of awareness among the adversarial machine learning community regarding the life cycle of a cybersecurity attack of traditional security threats and the interaction between a model and the system. While security defenses will ideally prevent all possible exploits against a system and its components, their practical goal is to collaborate with other defenses and hamper attacker efforts at every step in the cyberattack life cycle. The complex network of defenses may not prevent every attack, but they can provide time for a defender to react and remediate. Much of the community's focus has been on how to improve the adversarial robustness of models rather than the robustness of the system. Although, from a system security standpoint, state of the art evasion attacks have numerous flaws, (\eg highly visible reconnaissance) defenders are not allowed to exploit these flaws due to the commonly used ``adaptive attacker'' persona.

\section{Traditional Machine Learning Security}
\label{sec:background}
Machine learning (ML) systems have witnessed rapid adoption in diverse domains, driven by their ability to make data-driven predictions, automate tasks, and facilitate decision-making. However, the widespread deployment of ML models has raised significant concerns about their security and reliability.

\subsection{Vulnerabilities in Machine Learning Models}
Several classes of vulnerabilities associated with ML models jeopardize their deployment in real-world settings. These vulnerabilities can be broadly categorized into three main types: evasion, poisoning, and theft.
 
\subsubsection{Evasion Attacks}

Evasion attacks~\cite{szegedy2013intriguing,goodfellow2014explaining,papernot2017practical,carlini2017towards} occur when malicious actors manipulate input data to mislead ML models, causing them to make incorrect predictions or decisions. Adversaries aim to craft input samples that appear benign to humans but trigger worst-case behavior in the underlying ML model. There a two popular threat models for evasion attacks: (i)~\textit{white-box}: attackers have complete knowledge of the model architecture, its parameters, and the training data; (ii)~\textit{black-box}: attackers only have partial knowledge about the model and limited access to it, but have no knowledge about the its parameters or training data.

\subsubsection{Poisoning Attacks}

Poisoning attacks~\cite{biggio2012poisoning,chen2017targeted,shafahi2018poison,ye2019curiosity} involve adversaries injecting malicious data into the training dataset used to train ML models. These malicious data points are strategically inserted to compromise the model's integrity during training. In poisoning attacks, adversaries typically have access to the training data and aim to manipulate it so the trained model exhibits undesirable behavior when deployed. The threat model for poisoning attacks assumes that attackers can manipulate a portion of the training data.

\subsubsection{Model Theft Attacks}

Model theft attacks~\cite{tramer2016stealing,orekondy2019knockoff} revolve around the unauthorized extraction of ML model parameters or architecture. Attackers seek to reverse-engineer or replicate a target model by observing its predictions or other external behavior. Model theft attacks assume that adversaries have access to the model's output but not necessarily the model's training data or parameters.

\subsection{Evasion Attacks}

Evasion attacks were the first attacks discussed against machine learning models and have remained relevant even with the rise of large language models \cite{zou2023universal}. Given the pervasive use of ML models in decision-making systems, evasion attacks remain a popular threat in literature as exploitation enables adversaries to compromise the entire system for either profit or harm. For the remainder of the paper, we will focus our discussion on evasion attacks, but we note we are also critical of the feasibility of other published threats. Evasion attacks can be categorized into two primary types: (1) gradient-based attacks and (2) gradient-free attacks.

\subsubsection{Gradient-Based Attacks}

Gradient-based attacks~\cite{goodfellow2014explaining,madry2018towards,carlini2017towards} operate by leveraging gradients of the model with respect to a given loss function and input sample. White-box access to the target ML model is assumed in order to precisely compute the loss gradient, wherein the attacker possesses complete knowledge of the model's architecture and parameters. Attacks iteratively perturb the input sample in the direction that maximizes the loss while ensuring that the perturbations remain imperceptible. 

One of the most widely recognized gradient-based attacks is the Projected Gradient Descent (PGD) attack~\cite{madry2018towards}. This attack adopts an iterative approach to find the perturbation that maximizes a given loss (\eg cross-entropy loss). It begins with an initial randomly perturbed input and updates it iteratively by following the gradient direction of the model's loss function. The perturbed input is constrained within a predefined epsilon ball to ensure the changes remain imperceptible. Mathematically, the update step for PGD can be expressed as:

\[
x^{(t+1)} = \text{Clip}_{x, \epsilon} \left(x^{(t)} + \alpha \cdot \text{sign}(\nabla_x \mathcal{L}(x^{(t)}, y_{\text{true}}))\right)
\]

where \(x^{(t)}\) represents the perturbed input at iteration \(t\), \(\epsilon\) is the maximum perturbation allowed, \(\alpha\) is the step size, \(\nabla_x \mathcal{L}\) is the gradient of the loss with respect to input, and \(\text{Clip}_{x, \epsilon}\) constrains the perturbed input within the \(\epsilon\)-ball around the original input.

\subsubsection{Gradient-Free Attacks}
Gradient-free attacks~\cite{chen2020zoo,ilyas2018black,brendel2018decision,chen2020hopskipjump,andriushchenko2019square}, in contrast to gradient-based attacks, do not rely on gradient information and only need access to the prediction outputs. The two most common classes of gradient-free attacks are \textit{score-based attacks} and \textit{decision-based attacks}. Score-based attacks~\cite{ilyas2018black,chen2020zoo,andriushchenko2019square} assume the model's confidence scores are provided to optimize some objective function. Using a substantial number of queries, they approximate model information such as its loss gradient, or they can train a proxy model and perform a transfer attack. Square attack~\cite{andriushchenko2019square}, for example, proposed a query efficient $\ell_\infty$ and $\ell_2$ random search attack algorithm. It utilizes square shared noise and sampled noise distributions specifically crafted for $\ell_\infty$ and $\ell_2$ attacks. These distributions are motivated by the way images are processed by convolutional filters and the shape of the $\ell_p$-balls for different values of $p$. Decision-based attacks~\cite{brendel2018decision,chen2020hopskipjump}, on the other hand, rely only on the final decisions (\eg top-1). Attackers observe whether an input is classified as the target class and adjust the input accordingly based on a custom search scheme.


\subsection{How can we Prevent Evasion Attacks?}
Although numerous prior works have proposed mitigation or prevention techniques for adversarial evasion, Athalye~\etal demonstrated that most works unknowingly utilized \textit{gradient obfuscation}~\cite{athalye2018obfuscated}. A defense that prevents the model from generating useful gradients for an attack through shattering, stochasticity, or gradient vanishing/explosion is said to use gradient obfuscation. As they show in their paper, obfuscation does not measurably improve the adversarial accuracy of a defended model, as it can be easily bypassed by an attacker. As a result, adversarial training remains the simplest method to measurably improve the adversarial accuracy of a model.

\section{Examining the Feasibility of Adversarial Attacks}

The numerous examples of \textit{broken defenses}, \ie defense that can be broken by an adaptive attacker, highlighted by Athalye~\etal~\cite{athalye2018obfuscated} and others~\cite{carlini2017towards, CW_detect, adaptiveattacks} have caused the machine learning community to be skeptical about the usefulness of newly proposed adversarial defenses. This skepticism has raised the bar for publication; adversarial evasion defenses need to be almost perfect. Not only does a defense need to be secure against existing attacks for a given --often unrealistic-- threat model, but it must also remain secure against future attacks, or at least too complicated for a reviewer to identify flaws in the defense during the review process. Currently, identifying the flaws of a proposed defense against future attacks has been pushed to the authors of the proposal, but there is a problem here: reviewers err on the side of caution and often suspect insufficient evaluation by the authors of the proposal, even when no evidence exists. This suspicion often leads to rejection as additional evaluations that will satisfy the reviewers are unlikely to occur during the brief review window. This results in an equilibrium in which the authors are unmotivated to improve defenses. Furthermore, assuming that a defense paper is successfully published, future systematic evaluations of defenses by trained red teams could still undermine the accomplishment~\cite{athalye2018obfuscated}, further raising skepticism about these defenses.


Since adversarial defenses are put through such strict scrutiny, one would expect that adversarial attacks must be held to the same high standard. Interestingly, the evaluation of adversarial attacks has been more relaxed across the ML security community. About a decade ago, when adversarial attacks were first introduced, the community studied the nuances of adversarial attacks using simple image classification models under the white-box threat model~\cite{goodfellow2014explaining}. As most of the work was investigative and academic in nature, these papers only hinted at the \textit{possibility} of real-world exploitation without providing details as to how such an attack would occur in a real system. Although more recent works have diversified the machine learning applications they attack~\cite{roadsign,ehr_attack_usenix}, the attacks remain theoretical. Readers are cautioned against the possibility of exploitation, but the papers often rely on a black-box threat model and provide a vague description of a ``real attack'' that often does not hold up to scrutiny. For example, we present a high-level evaluation of an adversarial attack paper in USENIX 2023. We emphasize that this paper is not unique in its flaws but is a representative of the state of the art in the evaluation of what is considered to be ``real attacks''.

\subsection{\textit{How to Cover up Anomalous Accesses to Electronic Health Records}~\cite{ehr_attack_usenix}}
In this paper, the authors introduce an adversarial evasion attack against a machine learning system trained to detect inappropriate access to patient encounter information in a hospital. Given a set of access logs represented as a graph, the model must identify anomalous access requests (\ie edges between a requester and patient information) that require further review. The proposed adversarial evasion attack seeks to derive a series of access requests that are 1)~not flagged as anomalous by the detection model (covering access) and 2)~hide the access request to the target patient encounter information the attacker wants to retrieve (target access)).

\subsubsection{Threat Model}
It is assumed that the attacker, a hospital staff member such as a nurse or doctor, has white-box access to the detection model and has either complete or partial knowledge of the access logs. Partial knowledge of the access logs can be derived from the attacker's own patient encounters and the ones they observe by their colleagues. The attacker can make any access request (as a care provider) to any patient to mislead the detection model. 

\subsubsection{Issues for real-world exploitation}
Two issues with the proposed attack model hinder potential real-world exploitation:

First, white-box access to the detection model is a very strong assumption for a number of reasons. Although the authors justify that such access is possible because \textit{``...there are ways to get white-box model information via only black-box accesses''}, referring to methods that infer model attributes through query access, there are multiple issues with this argument in a real-world setting:
\begin{enumerate}
    \item \textbf{How does the hospital staff gain access to the anomaly detection model and/or its outputs?} - One would assume that the security aspects in the hospital, such as illegal data access requests, are handled by the hospital's IT staff rather than the doctors or nurses.
    \item \textbf{If such access was obtained, how did they hide their interactions with the model?} - In many adversarial machine learning attacks, the adversary is assumed to have unlimited query access to the model either through a public API or local copy. In both scenarios, interaction with the target model is necessary. In this particular attack, the attacker would need to request access to the model, an anomalous event in itself, and then bombard the model with spurious queries. 
\end{enumerate}

The second issue regards the attacker's knowledge of the access request graph. Even if an attacker has obtained white-box access to the detection model, complete knowledge of the access logs is unlikely to be known to hospital staff unless they are part of the IT staff. Thus, the attacker must rely on their own access requests and infer access requests of other staff members. In this scenario, as the authors demonstrate, the attack success rate with respect to the target access and covering access falls to 70\% and ~25\%, respectively, compared to ~77\% and 100\% when the entire graph can be observed. According to the authors, they post-process their data into 50 splits, where each split is meant to represent the access logs of a single week. A single data split has, on average, 82,498 accesses, and the detection model processes a single data split at a time. Thus, with respect to the author's dataset, 30\% of the graph represents ~24,749 access requests. 

According to a 2018 survey \cite{survey} of 8,774 physicians, a physician sees an average of 20 patients per day. This means that an attacker working seven days a week would only observe 140 access requests and need to accurately\footnote{The authors assume that an attacker has an accurate subgraph. There are no experimental results for when the subgraph is inaccurate.} derive an additional 24,649 requests. The authors provide no details on how accessible such information would be to the attacker.

\subsubsection{Takeaways}
Although the authors demonstrate an interesting theoretical attack on a healthcare detection model, it is unlikely for such an attack to be reliable against a real system. The authors make two strong assumptions, but, as discussed, these assumptions are likely unreasonable in a practical implementation of the attack. A real execution of the proposed adversarial attack would require the involvement of technical staff (\eg IT), but such involvement constitutes an insider threat, \ie the detection model can be bypassed entirely.

We reiterate that this evaluation is not meant to point out this particular paper as problematic but rather highlight the fact \textit{that adversarial attacks are evaluated less strictly than adversarial defenses}. Using the Athayle~\etal threat model, the assumptions made by this paper for a ``realistic'' attack scenario are quite generous as it greatly shortens the cyberattack lifecycle \cite{mitre:atlas}. This issue can be found in many other ``real'' attacks where the attacker has somehow gained extremely privileged access to the target model~\cite{roadsign}. Evaluating proposed attacks in such a relaxed manner is potentially harmful to the development of practical defenses as attacks do not consider the full capabilities of a defender but rather ignore potential security measures. Of course, it would be ideal if a model was perfectly secure in the absence of other protections, but we recognize security is created as a result of many interconnected defenses that specialize against certain types of threats.


\subsection{Adversarial Evasion in the Real World}
\label{sec:real_attacks}
Academic researchers have warned the community of the threat of adversarial attacks and encouraged the development of robust model defenses. Still, the unrealistic attack scenarios proposed in the existing literature make it hard for practitioners to accept. Recent surveys suggest that ML practitioners are either unaware or unconcerned with these attacks for a variety of reasons \cite{9283867, industryaml,grosse2022whydoso,bieringer2022industrial}. In fact, Bieringer~\etal~\cite{bieringer2022industrial} show that ML practitioners often do not have a clear mental model of adversarial ML. But, irrespective of industry beliefs, are there malicious actors using adversarial evasion, and if so, how? A recent work by Apruzzese~\etal~\cite{realattacks} studies the existence of such attacks on a real-world phishing webpage detection deployment. Their results show that while there were small traces of such adversarial inputs, none used any of the existing adversarial attacks. Instead, they relied on simple techniques such as blurring, cropping, and misspelling to avert the defense.  MITRE ATLAS has also documented several case studies either discovered in the wild or as a result of AI red teaming~\cite{mitre:atlas}. 

\paratitle{ClearviewAI Misconfiguration \cite{clearview}} Researchers at SpiderSilk found private source code and security credentials belonging to Clearview AI, a facial recognition company serving law enforcement and government agencies, that were publicly accessible due to a server misconfiguration. Using the leaked information, a malicious actor could download model information for use in an adversarial evasion attack, which would compromise existing and future AI systems deployed by Clearview AI.

This case study describes a white-box threat scenario with the potential for a black-box transferability attack. Due to a \textbf{data leak}, the attackers were able to directly extract the model and its information. Finally, as the attackers possess complete information about the model, they can deploy it \textbf{locally}, enabling \textbf{unlimited queries} to the model.

\paratitle{ProofPoint Evasion \cite{proofpoint}} Researchers at Silent Break Security discovered that Proofpoint, an email security company, was attaching its model outputs to the headers of emails scanned for spam. This data leak enabled the researchers to obtain a dataset of input/output pairs by sending a large number of emails to the service. They used this information to train a proxy model, through which they could attack using white-box methods and design adversarial examples to evade Proofpoint's detector model.

This case study describes a gray-box threat scenario. Due to a \textbf{data leak}, the model's exact outputs are provided, enabling a proxy model's training. Although the model exists in the victim's network, the model is part of an email scanning service, so it can be \textbf{freely queried}\footnote{By free, we refer to the number of queries that the service will respond to rather than any monetary cost.} 

\paratitle{Confusing Antimalware Neural Networks \cite{antimalware}} Researchers at Kaspersky performed a red team analysis of their malware detection model. Assuming a malware detection service hosted remotely in a cloud environment, a user can interact with the service and the back-end model by uploading information about a suspected malicious file. Due to potential legal restrictions and traffic limitations for uploading malicious files, they assume the user performs feature extraction locally before upload. Therefore, a malicious user can obtain a set of benign and malicious files, upload the features to the malware detection model, and label the files based on the model's response. Using this information, they can train a proxy model and adversarially attack it. The researchers demonstrated that this threat model created effective adversarial evasion samples.

This case study describes a gray-box threat scenario. Although the model exists in the victim's network, by design, the model can be \textbf{freely queried} and provides binary model feedback that can be used for training a proxy model.

\paratitle{Camera Hijack Attack on Facial Recognition System \cite{camerahijack}} In 2018, a pair of fraudsters exploited China's facial recognition system to defraud the tax system. Using high-definition photos of individuals that are easily accessible and a deepfake app, they created lifelike images, which were then passed directly to the facial recognition and liveliness detection model using a customized phone that bypassed the physical camera. This enabled the creation of synthetic identities for their shell company, which issued fake tax invoices. Although this incident was more of a brute-force attack, it represents a real case of adversarial evasion.

This case study describes a black-box threat scenario. The target model is contained in the app, which the attackers cannot access, but they can isolate it as it exists \textbf{locally}, \ie on the phone. Through camera hijacking, they can directly feed inputs into the model, and it appears the application allows for \textbf{unlimited queries}. This design allows them to brute-force attack the application without penalty.

From these case studies, a ``real'' adversarial attack appears to follow the following pattern: 
First, the adversary identifies a query feedback mechanism that allows it to perform an unlimited number of queries on the model. This can either happen through a query API that the attacker will use to train a proxy model or as part of a locally isolated model available to the attacker. Next, if the model is directly accessible, the attacker performs an adversarial evasion attack under a black/white-box setting depending on the scenario or otherwise attacks the proxy model using a white-box attack and attempts to transfer the successful samples.


A key step in this process, and something that previous works typically take for granted, is the first step. In traditional cybersecurity terms~\cite{mitre:attack}, this step is known as \textbf{reconnaissance}. The attacker is gathering information about the target system, namely the machine learning model to be evaded. If not inadvertently leaked, the attacker must rely on a query API to learn the input-output mappings of the model. Of the four case studies, three of them publicly exposed a query API by design. The last case study exposed the entire model, but the security issue was non-ML in nature. In the malware detection case study, the researchers explored possible adversarial countermeasures, but they only consider defenses that modify the victim model to improve performance on or detection of adversarial samples as is the norm in the community~\cite{antimalware}. All these attacks seem to implicitly allow malicious actors to freely probe models that are part of a service. In essence, reconnaissance is assumed to be easy as the defender  (\ie, the service provider) is making no attempts to prevent it. In the three case studies that expose a query API, there seems to be no discussion of non-ML countermeasures such as monitoring, logging, or user interaction analysis. 

We see this as a large blind spot in the ML security community and argue that we have been deluded by an unreasonable threat model that ignores traditional security measures.

\section{Traditional Solutions for New Problems}
\label{sec:results}
When a system is deployed publicly, there is an implicit understanding that not all of its interactions will be benign. A system should have measures in place to detect and respond to malicious activity. When suspicious activity is detected, the system can take steps to respond by enacting more vigorous defenses, raising an alert, and investigating the relevant activity logs. If the activity is confirmed to be malicious, remediation measures can be performed, such as terminating malicious connections, creating new security/detection rules, quarantining malicious files, etc. Generally, a deployed system is protected by both static and dynamic measures that monitor its state and react swiftly to threats. For example, Apruzzese~\etal~\cite{realattacks} looks at the ML-based spam detection system at Facebook and shows that the detection of spam content comes in after a long pipeline involving various traditional and ML-based defenses against automation, improper access, and malicious activity. 

While there has been a trickle of recent work that has started to look at such dynamic defenses in the context of stateful defense models~\cite{chen2020stateful,li2022blacklight}, the ``traditional'' adversarial attack threat model used in the community allows published attacks to ignore these measures. The system the model is part of is assumed to be isolated, static, or otherwise unable to respond. The defender only gets a single chance to deploy a defensive countermeasure. Therefore, any user interaction with the model is suspicious as one does not know if the input is benign or adversarial, hence the need for a robust model. It is true that if the attacker already has an adversarial input prepared, then a robust model is necessary, but what about the effort that goes into creating the input? As we observed in the case studies and related work, adversarial inputs are the product of multiple queries to the model. Therefore, before exploitation, \ie evasion, adversarial attacks must necessarily issue multiple \textbf{probing} queries. How visible is this reconnaissance, and what can be done to hinder or prevent it?



\subsection{The Visibility of Evasion}

To emphasize how visible adversarial evasion is during the probing phase, we trained a \resnets model \cite{he2015resnet} using standard cross-entropy loss (Natural) and adversarial training (Adversarial) on the \cifar dataset~\cite{krizhevsky2009cifar10}. For the white-box threat model, we used Projected Gradient Descent using an $\ell_{\infty}$-norm with an attack budget of $\epsilon = 8/255$ and step size of $T=2/255$~\cite{madry2018towards}. For the black-box threat model, we used the Square attack with the same parameters, an $\ell_{\infty}$-norm with an attack budget of $\epsilon = 8/255$~\cite{ACFH2020square}. We show the performance of these two pre-trained models against these attacks in Table~\ref{tab:train-acc}.

We attack each model using a random \cifar testing sample that was originally classified correctly; this is repeated for a total of 100 trials. The Adversarial Robustness Toolbox (ART), the library we used for the attacks, enables the generation of adversarial samples through the \texttt{generate()} function~\cite{art2018}. Each \texttt{generate} call can make several queries to the model. Therefore, we reported both the number of \texttt{generate} calls, \ie \textbf{rounds}, and the number of \textbf{queries} the model responded to. We present the results in Table~\ref{tab:attacker}.

For each model, in all 100 trials, both adversarial attacks succeeded in finding an adversarial sample. However, we note that regardless of the model or attack threat model, on average \textit{an adversarial attack needs to make multiple queries to the model before it is successful}. Improving the model's adversarial robustness with adversarial training increases the number of queries before a successful adversarial example is found. Limiting a user to black-box model access further increases the number of queries necessary, as the attacker must infer additional model information. 

\begin{table}[h!]
\centering
\caption{The accuracy of a pre-trained \resnets model evaluated against multiple adversarial evasion attacks. The model is trained on the \cifar dataset both naturally and adversarially. The clean accuracy of the model is indicated by ``none''.}
\resizebox{0.8\columnwidth}{!}{
\begin{tabular}{lcc}
    \toprule
    \textbf{Training Method} & \textbf{Attack} & \textbf{Accuracy} \\
    \midrule
    \multirow{3}{*}{Natural} & None & 93.07\% \\
    & PGD & 2.25\% \\
    & Square Attack & 42.67\% \\
    \midrule
    \multirow{3}{*}{Adversarial} & None & 82.60\% \\
    & PGD & 51.59\% \\
    & Square Attack & 81.32\% \\
    \bottomrule
\end{tabular}
}
\label{tab:train-acc}
\end{table}

\begin{table}[h!]
\centering
\caption{The simulation results of the attacker and defender interactions when running the PGD white-box and Square Attack black-box attacks. The simulation ends when misclassification occurs, and the attacker wins. The rounds and queries show the respective average among 100 trials along with the 95\% confidence. The results were generated using a ResNet-18 model.}
\resizebox{0.95\columnwidth}{!}{
\begin{tabular}{lccc}
    \toprule
    \multirow{2}{*}[-0.5em]{\textbf{Training}} & \multirow{2}{*}[-0.5em]{\textbf{Attack}} & \multicolumn{2}{c}{\textbf{Attacker Wins}} \\\cmidrule(lr){3-4}
    & & \textbf{Rounds} & \textbf{Queries} \\
    \midrule
    Natural & PGD  & $2.22 \pm 0.32$ & $6.66 \pm 0.97$ \\
    Adversarial & PGD & $6.94 \pm 0.84$ & $20.82 \pm 2.51$ \\
    Natural & Square Attack & $10.81 \pm 1.87$ & $63.35 \pm 11.27$ \\
    Adversarial & Square Attack & $15.34 \pm 2.49$ & $71.59 \pm 14.03$ \\
    \bottomrule
\end{tabular}
}
\label{tab:attacker}
\end{table}

Similar to observations made by recent stateful defense models~\cite{chen2020stateful,li2022blacklight}, we see that, when we examine the queries sent by both attacks, successive queries show a high degree of \textit{similarity}. White-box attacks issue queries that differ from the prior one by a small noise step along the adversarial loss gradient. Black-box attacks also issue additional estimation queries to derive an approximate loss gradient or identify the model's decision boundary. Thus, adversarial attacks appear to be highly visible, given the multiple highly similar queries issued in quick succession. In Section \ref{sec:discussion}, we provide a brief case study of ``normal'' query behavior.


\subsection{Detecting the Start of an Evasion Attack}

The traditional adversarial attack model ignores defensive measures that do not improve the adversarial performance of the model based on the reasoning that a knowledgeable attacker can bypass other defensive measures. But is this reasoning sufficient to allow attacks to interact with models so obviously? Consider a different security scenario, such as account authentication. Users need to provide security credentials to a system, and they are informed when authentication fails. In such systems, the number of failure attempts is limited to prevent brute force discovery of the correct credentials. If the limit is reached, the user is temporarily locked, and an alert is raised indicating suspicious activity. Current adversarial attacks are allowed to brute force attack the model due to the current threat model.

Let us study the implications of allowing an AI system to monitor and blacklist suspicious model interactions. The monitoring system does not need to identify if a given input would evade the model, nor does it need to filter out the effects of adversarial noise.  Rather, it just needs to evaluate if, based on the current query history, there is a possible adversarial attack occurring. This idea has recently gained traction in the literature under the umbrella of stateful defense models~\cite{chen2020stateful,li2022blacklight}. We borrow from prior work and implement the Blacklight detector in ART, a detection algorithm that raises an alert based on hash collisions using their similarity hash \cite{li2022blacklight}. We also implement an $\ell_{\infty}$ norm detector and a locality-sensitive hash detector that will raise alerts if there are queries whose $\ell_{\infty}$ norm is below a threshold of $8/255$, the same as the attack budget, or if a hash collision occurs, respectively. For all three detection techniques, we use a memory of the past 50 queries for detecting collisions. Now, in order for the attacker to win, they must both cause misclassification of the input and avoid detection. For simplicity, we will assume that the attacker still has white-/black-box query access and will differ the discussion regarding such access to Section \ref{sec:discussion}. 

\begin{table*}[h!]
\centering
\caption{The simulation results of the attacker and defender interactions when running the PGD white-box and Square Attack black-box attacks using various adversarial detectors. The attack wins when misclassification occurs, while the defender wins when it successfully detects the attacker is creating an adversarial example. The rounds and queries show the respective average among the attacker or defender victories along with the 95\% confidence. The results were generated using a ResNet-18 model.}
\resizebox{0.9\textwidth}{!}{
\begin{tabular}{lcccccccc}
    \toprule
    \multirow{2}{*}[-0.5em]{\textbf{Detector}} & \multirow{2}{*}[-0.5em]{\textbf{Training}} & \multirow{2}{*}[-0.5em]{\textbf{Attack}} & \multicolumn{3}{c}{\textbf{Attacker Wins}} & \multicolumn{3}{c}{\textbf{Defender Wins}} \\ \cmidrule(lr){4-6} \cmidrule(lr){7-9}
    & & & \textbf{Count} & \textbf{Rounds} & \textbf{Queries} & \textbf{Count} & \textbf{Rounds} & \textbf{Queries} \\
    \midrule
    \multirow{4}{*}{$\ell_{\infty}$ Norm} 
    & Natural & PGD & 35 & $1.00 \pm 0.00$ & $3.00 \pm 0.00$ & 65 & $2.00 \pm 0.00$ & $6.00 \pm 0.00$ \\
    & Adversarial & PGD & 6 & $1.00 \pm 0.00$ & $3.00 \pm 0.00$ & 94 & $2.00 \pm 0.00$ & $6.00 \pm 0.00$ \\
    & Natural & Square Attack & 34 & $3.27 \pm 0.53$ & $9.89 \pm 3.32$ & 66 & $3.61 \pm 0.51$ & $21.64 \pm 3.06$ \\
    & Adversarial & Square Attack & 24 & $5.58 \pm 1.34$ & $32.04 \pm 8.17$ & 76 & $5.60 \pm 0.89$ & $34.16 \pm 5.32$ \\
    \midrule
    \multirow{4}{*}{LSH} 
    & Natural & PGD & 49 & $1.11 \pm 0.11$ & $3.37 \pm 0.31$ & 51 & $2.16 \pm 0.50$ & $6.47 \pm 0.42$ \\
    & Adversarial & PGD & 13 & $2.46 \pm 1.06$ & $7.39 \pm 3.17$ & 87 & $2.63 \pm 0.27$ & $7.90 \pm 0.81$ \\
    & Natural & Square Attack & 17 & $2.53 \pm 0.81$ & $10.74 \pm 5.11$ & 83 & $2.91 \pm 0.27$ & $17.39 \pm 1.61$ \\
    & Adversarial & Square Attack & 8 & $2.67 \pm 1.03$ & $9.14 \pm 6.33$ & 92 & $3.59 \pm 0.52$ & $21.49 \pm 3.11$ \\
    \midrule
    \multirow{4}{*}{Blacklight} 
    & Natural & PGD & 41 & $1.07 \pm 0.10$ & $3.22 \pm 0.31$ & 59 & $2.07 \pm 0.06$ & $6.20 \pm 0.19$ \\
    & Adversarial & PGD & 7 & $1.29 \pm 0.34$ & $3.86 \pm 1.01$ & 93 & $2.07 \pm 0.05$ & $6.19 \pm 0.15$ \\
    & Natural & Square Attack & 24 & $2.22 \pm 0.47$ & $11.63 \pm 2.76$ & 76 & $2.86 \pm 0.23$ & $17.11 \pm 1.42$ \\
    & Adversarial & Square Attack & 13 & $3.33 \pm 1.68$ & $18.08 \pm 10.19$ & 87 & $3.73 \pm 0.49$ & $22.36 \pm 2.94$ \\
    \bottomrule
\end{tabular}
}
\label{tab:detector}
\end{table*}

As Table~\ref{tab:detector} demonstrates, the obvious nature of traditional adversarial attacks is easily caught by our detectors \footnote{Note that the number of queries will be higher than the number of rounds as our detectors operate on the output of a generate call and each generate call can issue multiple queries.}. Unless the PGD attacker wins in a single round, \ie the first adversarial noise step creates an adversarial sample, they are immediately caught upon issuing their next query most of the time. The Square attack attacker issues slightly more queries as there is more diversity in their behavior due to the estimation queries, but detection remains effective.

\subsection{What About Evasion Strategies?}
In the previous section, we showed the effectiveness of simple detection techniques in thwarting adversarial attacks. The obvious question to raise here is, ``What about adaptive attackers?''. The answer to this question often breaks many proposed defenses, but accurately obtaining the defense information is not as trivial as the community might believe. Reconnaissance is very demanding for cybersecurity attacks, especially when targeting proprietary systems. However, published attacks gloss over both the specific and high-level details of how to perform this. Nevertheless, it is reasonable (and should probably always be assumed) for the attacker to suspect that some form of detection is occurring on the model inputs. To evade detection by Blacklight, we modify the attacks so they will issue \textit{benign} queries for a number of rounds in between adversarial rounds. Benign queries are drawn randomly without replacement from a set of test inputs never seen by the detector. The attack does not retain the information returned by a benign query. We use the adversarially trained ResNet-18 model. 

\begin{table*}[h!]
\centering
\caption{The simulation results of the attacker and defender interactions when running the PGD white-box and Square Attack black-box attacks using an adversarially trained ResNet-18 defense model and the Blacklight detector where the attack is capable of evasion. The attacker evades by performing a no-op attack by sampling a random image from the \cifar dataset for a certain amount of turns. The attack wins when misclassification occurs, or the defender issues a false detection. The defender wins when it successfully detects the attacker has issued a non-benign query. The rounds and queries show the respective average among the attacker or defender victories along with the 95\% confidence. The results were generated using a ResNet-18 model.}
\resizebox{0.9\textwidth}{!}{
\begin{tabular}{lccccccc}
    \toprule
    \multirow{2}{*}[-0.5em]{\textbf{Attack}} & \multirow{2}{*}[-0.5em]{\textbf{Evasion}} & \multicolumn{3}{c}{\textbf{Attacker Wins}} & \multicolumn{3}{c}{\textbf{Defender Wins}} \\ \cmidrule(lr){3-5} \cmidrule(lr){6-8}
    & & \textbf{Count} & \textbf{Rounds} & \textbf{Queries} & \textbf{Count} & \textbf{Rounds} & \textbf{Queries} \\
    \midrule
     \multirow{4}{*}{PGD}
     & 0 Rounds  &  7 & $1.29 \pm 0.33$    & $3.86 \pm 1.00$   & 93 & $2.06 \pm 0.05$    & $6.19 \pm 0.15$  \\
     & 7 Rounds  &  14 & $8.57 \pm 1.08$    & $3.21 \pm 0.40$   & 86 & $17.12 \pm 0.64$   & $6.42 \pm 0.24$  \\
     & 15 Rounds &  20 & $19.20 \pm 2.80$   & $3.60 \pm 0.53$   & 80 & $34.11 \pm 1.66$   & $6.40 \pm 0.31$  \\
     & 35 Rounds &  25 & $39.60 \pm 4.23$   & $3.30 \pm 0.35$   & 75 & $77.14 \pm 1.46$   & $6.43 \pm 0.28$  \\
     \midrule
     \multirow{4}{*}{Square Attack}
     & 0 Rounds  &  13 & $3.33 \pm 1.68$    & $18.08 \pm 10.19$ & 87 & $3.73 \pm 0.49$    & $22.36 \pm 2.94$ \\
     & 7 Rounds  &  16 & $27.50 \pm 12.55$  & $19.06 \pm 9.36$  & 84 & $28.76 \pm 3.89$   & $21.55 \pm 2.89$ \\
     & 15 Rounds &  19 & $59.08 \pm 25.92$  & $20.61 \pm 9.61$  & 81 & $63.08 \pm 7.11$   & $23.61 \pm 2.67$ \\
     & 35 Rounds &  23 & $109.73 \pm 22.27$ & $16.55 \pm 3.59$  & 67 & $126.39 \pm 16.19$ & $21.10 \pm 2.71$ \\
    \bottomrule
\end{tabular}
}
\label{tab:evasion}
\end{table*}

As we see in Table~\ref{tab:evasion}, when the number of evasion rounds is less than or equal to the detector's cache window, 50 samples, the attacker is detected in the next adversarial round. For example, with 7 rounds of benign queries, the PGD attacker only wins on average if their adversarial query in round 8 succeeds; otherwise, they are often caught the next time they query in round 16. When the attacker uses an evasion window larger than 35, their attack success rate is almost 100\%. Therefore, it is possible for an attacker to evade detection with a simple strategy, but there are two caveats: First, to evade detection, they must issue additional queries, which can impose temporal or financial costs on the attacker. Debenedetti~\etal~\cite{debenedetti2023evading} has recently explored this challenge in the context of black-box evasion attacks where they sought to minimize the number of ``bad'' queries to a classifier. Second, they need to correctly guess the detection cache window before attacking. If they underestimate the window, they will be detected and alert the system, allowing it to deploy countermeasures such as temporarily increasing the sensitivity of detection systems, more aggressive rate limiting, or blacklisting. If they overestimate by too much, they will pay additional costs as before. These principles similarly apply to more sophisticated attacks, such as a recently proposed one by Fend~\etal~\cite{feng2023stateful} against stateful defenses. In any case, even if detection does not stop adversarial attacks, it slows them down, providing the system time to respond.

\section{Exploring Other Attack Assumptions} \label{sec:discussion}
While we have seen many different proposals for how to attack a machine learning system, there has been little exploration using traditional security principles in how to defend them. In the previous section, we explored one way of defending an AI system by monitoring the query API and blacklisting potential attackers. In our exploration, however, we made some simplifying assumptions, which we will discuss now.

\subsection{Multi-user Scenarios}
Our experiments were performed under the premise of a single adversarial user interacting with the model. Therefore, the detection of adversarial probing queries is greatly simplified as the query history cache only needs to store and compare data for a single user. In reality, an AI system will likely serve multiple users concurrently, complicating the detection of adversarial probing queries. While we recognize that an increased user count will affect the cache size and benefit the attacker, the main new challenge is the attacker collaboration. Suppose that for each user, the detector tracks a separate cache history. One or more attackers could create multiple user accounts and have each account issue one adversarial query to avoid collisions in the user cache. To avoid this evasion strategy, the detector must compare a user's query to the history of all other user queries. But would this multi-user detection approach result in numerous false positives?

We explored this scenario using two vision and one speech dataset. For this experiment, we run the Blacklight detector on $10,000$ images from \cifar, $117,577$ images from the Landmarks~\cite{weyand2020GLDv2}, and $4,890$ voice command from the Speech Commands dataset~\cite{warden2018speech} constituting their respective test sets. Table~\ref{tab:collisions} summarizes these results. In all three cases, our experiments caused zero false-positive collisions between the samples. This suggests that detection may be appropriate even for multi-user environments where data samples may appear more similar (\eg picture from a similar landmark or different users providing the same voice command). Although there may very well be scenarios where the use of such detectors is infeasible, we encourage practitioners to examine such ML defenses from multiple perspectives and explore their applicability on their end-to-end system. 





\begin{table}[h!]
\centering
\caption{The number of false-positive collisions that occurred from running Blacklight on the test set of various datasets.}
\resizebox{0.8\columnwidth}{!}{
\begin{tabular}{lccc}
\toprule
\textbf{Dataset} & \textbf{Samples} & \textbf{Classes} & \textbf{Collisions} \\
\midrule
CIFAR-10        & 10,000  & 10     & 0   \\
Speech Commands & 4,890   & 12     & 0   \\
Landmarks       & 117,577 & 12,894 & 0   \\
\bottomrule
\end{tabular}
}
\label{tab:collisions}
\end{table}

\subsection{Explicit Feedback Assumption}
Another simplifying assumption we made in the previous section was that attackers had access to an explicit query feedback mechanism that provided them with either white-box or black-box information necessary for the attack. Obviously, a query API providing white-box information is highly unlikely, but what about black-box information? In Section \ref{sec:real_attacks}, we observed that in the four ATLAS case studies, three of the applications served users by providing a classification result for a given input. Thus, providing black-box information was necessary. In such scenarios, where a query API must necessarily be provided, the black-box threat model is reasonable. Our results in Section \ref{sec:results} suggest that monitoring can only slow down evasion if the attacker is actively probing the model. An attacker could also just issue benign queries to train a proxy model and attack that model offline. We do not explore potential solutions to model theft in this paper.

However, machine learning is used in many applications, and not all require exposing a public query API. The work by Xu~\etal mentioned earlier shows how hospitals use machine learning models to detect anomalous record access. We also know that enterprises use machine learning models to detect anomalies and malware in the network. In such cases, these systems do not, and should not, expose a public API to query the model and obtain feedback. Obtaining feedback in such cases is a non-trivial problem that prior attack works ignore. Is explicit feedback possible in these scenarios, or can attackers only rely on indirect feedback, \eg the malware stops beaconing, indicating it was detected? What is the additional temporal cost of indirect feedback? 
We recommend that new attacks include at least some indication of the real requirements necessary for exploitation so defenders can identify potential opportunities to protect AI systems.

\subsection{Asymmetric Cost of Queries}
Another simplifying assumption frequently made by attackers is to assign equal cost to positive and negative queries. In many security-sensitive ML applications, such as anomaly detection, intrusion detection, and malicious content detection, however, there can be significant consequences for triggering the system where additional security mechanisms such as rate-limiting, block listing, and human intervention may be triggered upon arrival of a ``bad query''.
There have been recent efforts to understand the challenges and tradeoffs of such queries. Most notably, Debenedetti~\etal~\cite{debenedetti2023evading} highlighted the asymmetric cost of bad queries and devised new ``stealthy'' versions of existing attacks to minimize the number of bad queries issued to a few hundred. While this is a promising first step toward designing stealthy adversarial attacks, even a few hundred bad queries are plenty to alert defenders to suspicious activity. Their work further highlights the challenges of realizing realistic attacks across ML-powered systems that rely on non-ML-based security measures to increase the barrier of entry for attackers. 

\subsection{Beyond Evasion Attacks}
We focused on adversarial evasion attacks given their prevalence in the literature, but as we discussed in Section \ref{sec:background}, there are many other threats to AI. The industry has shown increasing concern with threats such as model theft and adversarial poisoning due to the potential monetary impact and negative publicity that would result from an AI incident~\cite{survey}. Our message remains the same: Analyze the attack lifecycle of the threat and identify potential mitigation opportunities across the AI pipeline. Study traditional security defense techniques as the model is not the only place where mitigations can be implemented.

For example, most adversarial poisoning defenses focus on one of two approaches: 1) Detect and filter out poisoned samples in the training data or 2) Design a robust training scheme to reduce the impact of poison samples. Both approaches try to remove the impact of poisoned samples but can have side effects such as hurting performance on minority sub-populations in the data \cite{baracaldo2022benchmarking}. In 2022, Shan~\etal proposed a novel approach to mitigating poison attacks based on digital forensics~\cite{281308}. Rather than preventing the poisoning of the model, they improved the recovery and remediation process after a poisoning incident occurred. In other security domains, recovery and remediation is a critical step, as it can be difficult to catch every single attack, especially 0-day attacks, but it was absent in the AI field until this work.

\section{Discussion}
Some may argue that there was nothing wrong with the state of adversarial machine learning: defenses have always needed to satisfy a ``for all...'' constraint, whereas attacks need only satisfy a ``there exists...'' constraint. We, and others~\cite{grosse2024}, respectfully disagree as there has been insufficient consideration with respect to the real world problem constraints and realistic threat models. We worry that the early failures at securing machine learning models have scared the community into being overly distrustful of any new attempts. A defense could be and has been rejected due to unrealistic attack suspicions, \eg adaptive attacks, without a clear attack in mind. This conservative mindset may discourage authors from exploring future work.

In ICLR 2020, a defense proposal was rejected \cite{rejected1} because the reviewers ``\textit{think that this paper is a simple extension of gradient masking...}''. Specifically, reviewer 3 mentions that ``\textit{[they] still think the algorithm in this paper is merely a clever version of gradient masking, which does not give the neural networks real robustness, it is just harder to design attacks on all these discrete operations}''. In a different rejected paper submitted to the same conference \cite{rejected2}, reviewer 2 suggested the authors follow community evaluation guidelines, including performing adaptive evaluation. The authors had already done so and asked how to design a more effective adaptive attack, but received no response from the review committee. We emphasize that we are neither affiliated with these parties nor make any judgment regarding the effectiveness of the proposed defenses. However, as demonstrated in the reviews and discussions, “for all…” is difficult to prove, especially when not confined within a realistic threat model. Looking at evasion-focused papers accepted at IEEE S\&P and Usenix Security in 2023, two major security conferences, we see 21 attack papers vs. only 4 defense papers.

We recognize that the enormous number of paper submissions makes it hard to identify promising defenses, especially because reviewers cannot empirically verify the results. Furthermore, the community should not publicize papers that provide false security promises. These issues could be alleviated with community guidelines regarding formal threat models and acceptable evaluation criteria, but only through publicizing defenses evaluated against realistic threat models can we foster healthy community discussion and encourage innovation. AI-Guardian~\cite{AIGuardian} was improved after publication, despite its flaws, due to penetration testers and authors working together to clearly identify the risks and opportunities \cite{AIGuardian-improve}. As generative AI applications become more widespread, we need potential countermeasures that the community can improve upon.

\section{Conclusion}
The field of adversarial machine learning appears to be at an impasse. While new attacks on AI systems continue to emerge at top-tier conferences, we see limited advancements in defenses. Adversarial evasion attacks have been studied for about a decade, but we, as a community, still seem to lack a unified stance on how to mitigate the threat. Our best advice appears to be \textbf{train more}, but that is unsatisfying for practical deployments that have many real-world deployment concerns, especially as models get larger. Although improving model robustness is one potential mitigation, prior work in adversarial AI appeared overly focused on it due to a flawed threat model. Machine learning models are part of a \textit{system} and should be defended as such. Systems are not defended by a single component but by layers of components that work together to harden exploitation and address a wide range of threats.

This paper is not trying to diminish the contributions of prior work but rather serves as a warning for the future. Generative AI has dramatically reshaped the technological landscape, but some of the new AI problems it brings appear to be similar to the AI problems of the previous decade. Large language models are all vulnerable to the new face of adversarial evasion, \ie \textit{prompt injection} and \textit{jailbreaking}~\cite{zou2023universal, liu2024autodan}. Various post- and pre-processing methods have been developed but are discovered to be bypassable~\cite{llamajailbreak}. \textit{Safety alignment}, a.k.a. adversarial training, adds relabeled attack samples to the training data to shrink the attack surface of large models. We worry that generative AI research will follow a similar pattern as the decade of research that came before it.

\bibliographystyle{plain}
\bibliography{main}

\end{document}

%% file: ethos.tex
\usepackage{color}
\usepackage{xspace}
\usepackage{url}
\usepackage{hyperref}
\usepackage[font={footnotesize},labelfont=bf]{caption,subfig} 
\usepackage{pifont} 
\usepackage{amsmath}

\usepackage{silence} 
\usepackage[normalem]{ulem} 

\usepackage{enumitem} 
\setlist{nosep} 
\setlist[itemize]{leftmargin=*} 

\usepackage{wrapfig}
\usepackage{rotating} 

\usepackage{booktabs} 

\usepackage{array} 
\newcommand{\PreserveBackslash}[1]{\let\temp=\\#1\let\\=\temp}
\newcolumntype{C}[1]{>{\PreserveBackslash\centering}p{#1}}
\newcolumntype{R}[1]{>{\PreserveBackslash\raggedleft}p{#1}}
\newcolumntype{L}[1]{>{\PreserveBackslash\raggedright}p{#1}}

\usepackage[compact]{titlesec}

\newcommand{\cifar}[0]{CIFAR-10\xspace}

\newcommand{\resnets}[0]{ResNet-18\xspace}

\newif\ifsubmit
\submitfalse

\ifsubmit
    \newcommand{\amir}[1]{}
    \newcommand{\kevin}[1]{}
    
    \newcommand{\todo}[1]{}
    \newcommand{\tone}[1]{}
\else
    
    \newcommand{\amir}[1]{\textcolor{blue}{Amir: #1}}
    \newcommand{\kevin}[1]{\textcolor{green}{Kevin: #1}}
    
    \newcommand{\todo}[1]{\textcolor{red}{TODO: #1}}
    \newcommand{\tone}[1]{\textcolor{orange}{[TONE] #1}}
\fi

\newcommand{\paratitle}[1]{\noindent\textbf{\textit{#1.}}\xspace}

\newcommand{\ie}[0]{\emph{i.e.,}\xspace}
\newcommand{\etal}[0]{\emph{et~al.}\xspace}
\newcommand{\eg}[0]{\emph{e.g.,}\xspace}

\usepackage{cleveref} 

\crefformat{section}{\S#2#1#3} 
\crefformat{subsection}{\S#2#1#3}
\crefformat{subsubsection}{\S#2#1#3}